\documentclass[11pt]{article}

\usepackage{PRIMEarxiv}

\usepackage[utf8]{inputenc} 
\usepackage[T1]{fontenc}    
\usepackage{hyperref}       
\usepackage{url}            
\usepackage{booktabs}       
\usepackage{amsfonts}       
\usepackage{nicefrac}       
\usepackage{microtype}      
\usepackage{xcolor}         
\usepackage{lipsum}
\usepackage{fancyhdr}       
\usepackage{graphicx}       
\usepackage{amsmath}
\usepackage{comment}
\usepackage{subcaption}
\usepackage{enumitem}
\usepackage{cleveref}
\usepackage{array}
\usepackage[ruled,longend,linesnumbered]{algorithm2e}
\usepackage[flushleft]{threeparttable}
\usepackage{bbm}
\usepackage[numbers]{natbib}

\title{Survey on Memory-Augmented Neural Networks: Cognitive Insights to AI Applications}

\author{
    Savya Khosla\thanks{Equal contribution. Savya Khosla wrote the section on memory-inspired neural networks, Zhen Zhu wrote the section on memory threories, and Yifei He wrote the secion on applications of memory-augmented neural networks and the discussion on the future research directions.} \qquad \qquad
    Zhen Zhu\footnotemark[1] \qquad \qquad
    Yifei He\footnotemark[1] \\ \\
    Universiy of Illinois Urbana-Champaign \\
    \tt\small \{savyak2, zhenzhu4, yifeihe3\}@illinois.edu
}

\begin{document}
\maketitle

\begin{figure}[!h]
    \centering
    \includegraphics[width=.85\linewidth]{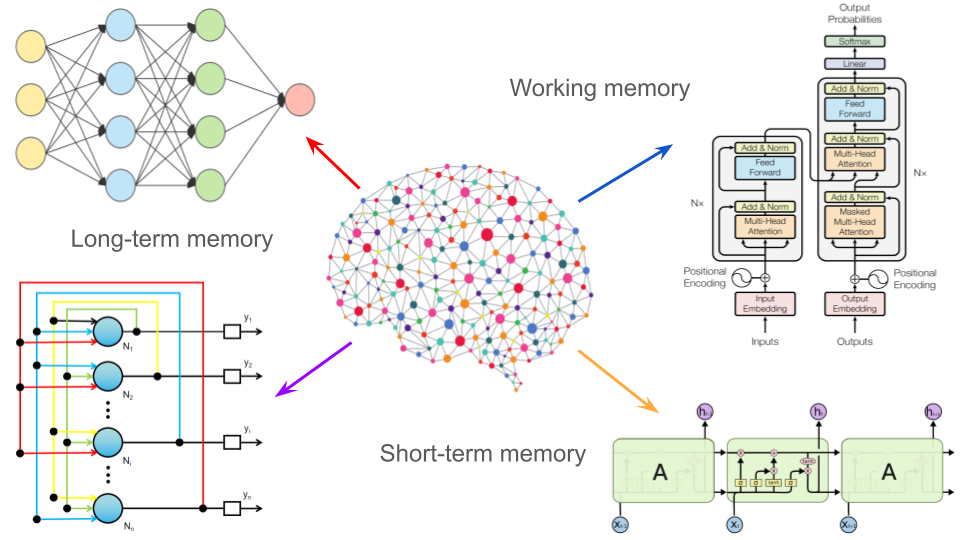}
    \label{fig:teaser}
\end{figure}

\vspace{1cm}

\begin{abstract}
    This paper explores Memory-Augmented Neural Networks (MANNs), delving into how they blend human-like memory processes into AI. It covers different memory types, like sensory, short-term, and long-term memory, linking psychological theories with AI applications. The study investigates advanced architectures such as Hopfield Networks, Neural Turing Machines, Correlation Matrix Memories, Memformer, and Neural Attention Memory, explaining how they work and where they excel. It dives into real-world uses of MANNs across Natural Language Processing, Computer Vision, Multimodal Learning, and Retrieval Models, showing how memory boosters enhance accuracy, efficiency, and reliability in AI tasks. Overall, this survey provides a comprehensive view of MANNs, offering insights for future research in memory-based AI systems.
\end{abstract}
\section{Introduction}
Memory-augmented neural networks have emerged as a powerful paradigm for addressing complex tasks that require the ability to store and retrieve information over extended sequences. Normal neural networks are ineffective in learning emergent concepts, adapting to new environment, dealing with a different task. Built-in memory mechanisms enable aggregating relevant information instantly, thus being much powerful in dealing with in-context tasks. This survey aims to conduct an analysis of the existing literature and advancements in memory-augmented neural networks, with the goal of providing a clear and in-depth understanding of this field.

The survey is comprised of three major topics.

\begin{enumerate}
    \item Memory theories: We will explore different types of memory mechanisms and supporting materials, trying to draw connections between memory theories from psychology to applications in AI.
    \item Architectures: We will investigate and categorize the various architectures of memory-augmented neural networks like Hopfield networks \cite{hopfield1982neural}, Neural Turing Machines (NTMs) \cite{graves2014neural}, and Transformer-based models with memory components \cite{wu2022memformer, katharopoulos2020transformers}. We plan to delve into the design principles, structures, and mechanisms of these models.
    \item Applications: Our survey will cover a wide range of applications where memory-augmented neural networks have been successfully applied. This includes natural language processing \cite{borgeaud2022improving, lewis2021retrievalaugmented}, computer vision \cite{long2022retrieval, blattmann2022semiparametric}, multimodal learning \cite{chen2022murag}, and other domains.
\end{enumerate}

Memory-augmented neural networks have the potential to revolutionize various AI applications by enhancing their ability to store data, reason and generalize. However, to the best of our knowledge, there is no comprehensive survey consolidating the different techniques of augmented neural networks with memory. So, through this survey we aim to consolidate the existing knowledge in this area, making it accessible and informative for both newcomers and experts.
\section{Memory Theories: From Psychology to Artificial Intelligence}
\label{sec:memory_theories}

Memory is a fundamental aspect of human cognition, and it plays an equally crucial role in the world of computer science and artificial intelligence. This section of the report, delves into the diverse memory mechanisms that underpin our cognitive processes and investigates how these concepts can be translated into the realm of AI. We'll navigate through various memory theories from the field of psychology, seeking to establish connections on how these theories have inspired several methods in artificial intelligence. 


\subsection{Memory Process and Storage}

\begin{figure}[!h]
    \centering
    \includegraphics[width=0.8\linewidth]{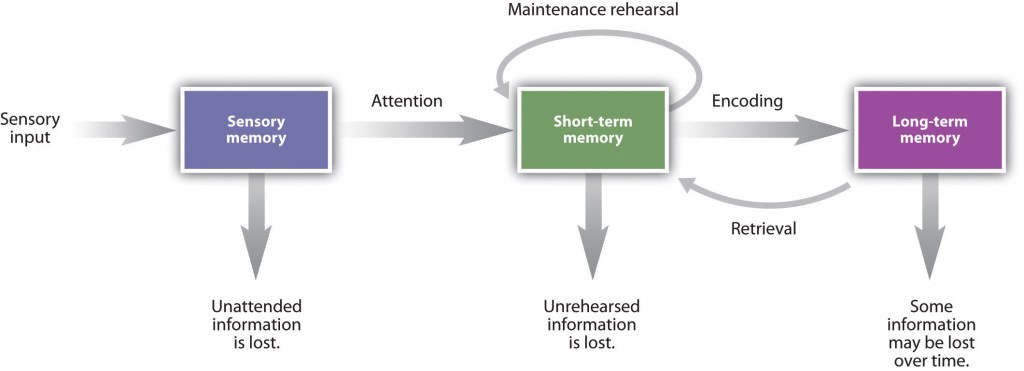}
    \caption{Memory process. Figure is taken from~\cite{stangor2014introduction}.}
    \label{fig:memory_process}
\end{figure}

According to the Atkinson-Shiffrin model~\cite{atkinson1968human}, memory is generally understood as an information process that incorporates sensory memory, short-term memory, and long-term memory~\cite{baddeley2007working}. As shown in Fig.~\ref{fig:memory_process}, information is received by sensory receptors and transformed into sensory input to the sensory memory. Much of the information is lost if the attention system does not attend to. The attended information would then flow to the short-term memory but without rehearsal, it still gets filtered and the rehearsed memory has the chance to be encoded to the long-term memory. But even the long-term memory is not a permanent store. 

This information distillation process is vital to our survival, even assuming the brain has infinite capacity to store every piece of information in the memory storage. On the one hand, the process allows us to prioritize, group, and generalize on processed information. On the other hand, filtered information may be useless or even harmful to our existence, such as traumatic experience. 

\subsection{Sensory Memory}

Sensory memory is a small buffer that holds sensory impressions from sight, hearing, smell, taste, and touch, for less than a second.  The sensory memory allows the brain time to process information in a continuous manner, otherwise we might see the world as discrete pieces rather than an unbroken stream of events~\cite{stangor2014introduction}. Though the storage is small and the duration is short, it is still important to our survival. For example, without sensory memory, our motor skills such as driving and skateboarding, would be largely impaired due to the lack of sensing the world in a continuous manner. 

\subsection{Short-term Memory}
\label{stm}

Short-term memory is temporary storage that holds a small amount of information in an active, readily available state for a short interval \footnote{\href{https://en.wikipedia.org/wiki/Short-term_memory}{Wikipedia of short-term memory}}. An example of short-term memory is the working desk: you place items on the desk for your current task; once the work is done, those items are either put away to another place or discarded permanently, to make room for new tasks. 

The duration of the short-term memory is usually a few seconds to a minute. Though still short in duration, the short-term memory is necessary for many daily tasks such as temporally memorizing a phone number, an address, or conversations. Without it, our daily life would be significantly hampered. The early symptoms of Alzheimer's disease is the loss of short-term memory~\footnote{\href{https://thekensingtonredondobeach.com/short-term-memory-loss-sign-of-alzheimers-or-dementia/}{Short term memory loss: sign of Alzheimer’s or dementia?}}

According to Georage Miller's research~\cite{miller1956magical} on the capacity of the short-term memory, short-term memory puts information in chunks and the average capacity is seven. Since the capacity can vary case by case, so he further appended ±2 items on 7. Though the actual capacity is not deterministic to each person, the idea is clear that the short-term memory has a limited capacity.

Working memory and short-term memory are usually used alternately from a historical reason. The concept of working memory emerged with the study of short-term memory. The understanding of short-term memory evolved from a unified storage system to functional usages as manipulation and information processing. Eventually, it led to the study of Baddeley and Hitch in 1974~\cite{BADDELEY197447} which brought up working memory to distinguish from the short-term memory.
Though definition differs, our interpretation for the working memory aligns with~\cite{cowan2008differences}: working memory is a part of the short-term memory but is mainly responsible for directing the attention to manage short-term memory. 

\subsection{Long-term Memory}

\begin{figure}[!h]
    \centering
    \includegraphics[width=0.5\linewidth]{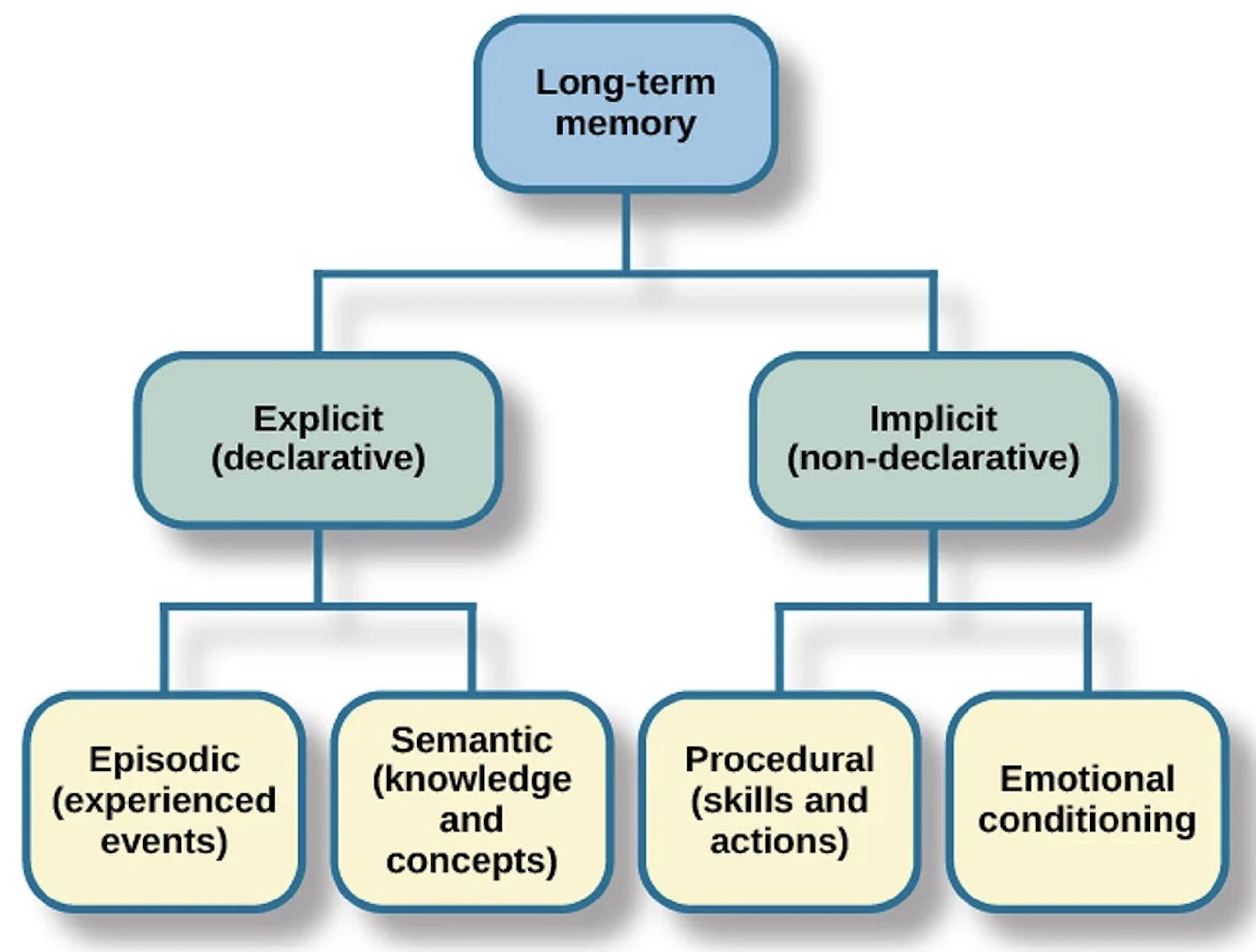}
    \caption{Memory process. Figure is taken from~\href{https://www.simplypsychology.org/long-term-memory.html}{Long-Term Memory In Psychology: Types, Capacity \& Duration}.}
    \label{fig:long_term_memory}
\end{figure}

The next stop for the information processed by the short-term memory is the long-term memory. Unlike the short-term memory which is only able to hold a few items in a short period of time, long-term memory can hold much more information for a much longer time. The duration of long-term memory is usually believed to reach years, or even lifetime. Examples of long-term memory are the recollection of an important event in distant past or bicycle riding skills someone learned in childhood. Fig.~\ref{fig:long_term_memory} depicts clear categorization of long-term memory. Episodic memory stores life events that happened in the past. Semantic memory contains factual information such as the the population of the US. and the meaning of words. Procedural memory generally deposits skills such as the motor skills for driving and swimming and is responsible for knowing how to do things. Emotional memory colors life events with emotions and therefore strengthen memorization of certain information. These memories collaboratively defines identities of individuals and crucial for the existence of individual entities. Once formed, long-term memories are less prone to fade or intervened compared to short-term memory. This explains why after years of not riding after the learning how to ride, a person is able to ride a bike easily since the riding skills are stored in the long-term memory and don't fade for long.

Long-term memory largely exists in the synapses of neurons distributed over cortex. It is analogous to artificial neural networks which stores information in the weights of neural connections. However, unlike the stablility of long-term memory in memory interference, artificial neural networks are prone to ``catastrophic forgetting''~\cite{ratcliff1990connectionist}. To figure out how to prevent forgetting necessitates understanding of the memory consolidation process.



\subsection{Memory Consolidation}

Memory consolidation is a critical process that stabilizes memory traces after their initial formation, primarily during restful states like sleep. This phase involves the brain reactivating and reorganizing memories, weaving them into existing knowledge networks~\cite{frontiers2013}. 
Though initially fragile and prone to disruption, memories become more stable and resistant to decay and interference through consolidation, significantly lowering the likelihood of forgetting. Essentially, the more robust the memory trace, the less likely it is to fade.
Synaptic plasticity, the ability to modify synaptic strength, supports these processes. Mechanisms like long-term potentiation (LTP) reinforce synaptic connections through repeated activation, making memories stronger each time they are accessed. Practices such as regular rehearsal enhance memory retention, countering our natural inclination to forget~\cite{goto2022offline}.

Besides fortifying memory, memory consolidation also rearranges them. This reorganization leads to a more efficient allocation of memories across various brain regions. Initially dependent on the hippocampus, over time, memories are increasingly stored in the neocortex, enhancing their durability and resistance to forgetting.

However, memory consolidation isn't foolproof against forgetting. Factors like new learning can disrupt the consolidation of older memories (retroactive interference), and previously consolidated memories can hinder the consolidation of new information (proactive interference).

\section{Memory-Inspired Neural Architectures}

In this section, we will discuss some memory-inspired neural networks. In particular, we elaborate on how models like recurrent neural networks and Transformers mimic specific types of human memory mechanisms, and we discuss some architectures designed to capture long-term memory - Hopfield Networks, Correlation Matrix Memories, Neural Turing Machines, Memformer, and Neural Attention Memory.

\subsection{Recurrent Neural Networks (RNNs)}
RNNs function much like our short-term memory. They retain recent information using a hidden state, akin to recalling something from just a moment ago. These RNNs excel in understanding patterns by tracking past elements in a sequence to anticipate what might follow. However, like our short-term memory that weakens over time, vanilla RNNs tend to forget older details as they process new data.

LSTMs, or Long-Short Term Memory networks, \cite{sepp1997lstm} are designed with the objective of remembering both recent stuff and things from a while back, just like our short-term and long-term memory. The hidden part in LSTMs is like the short-term memory, holding onto what just happened, while the cell state acts like the long-term memory, keeping important stuff for a longer time. These two parts work together, deciding what to remember now and what to keep for later, just like human brain sorts out what is important in the moment and what is worth remembering in the long run. This way, LSTMs mimic human brain's way of handling short-term and long-term memories.

\begin{figure}[!h]
    \centering
    \includegraphics[width=.95\linewidth]{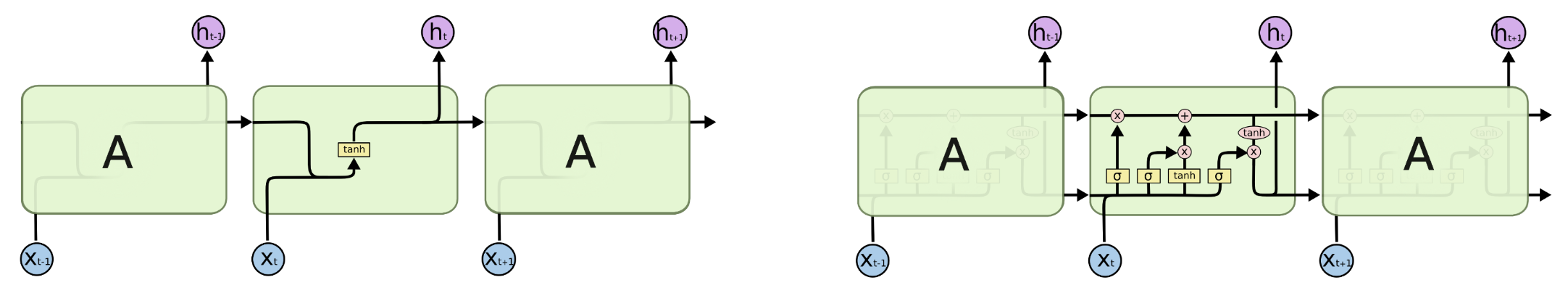}
    \caption{The hidden state of RNNs is similar to short-term memory and the cell state of LSTMs is similar to long-term memory. Figure taken from \href{https://colah.github.io/posts/2015-08-Understanding-LSTMs/}{Understanding LSTM Networks by Christopher Olah}}
    \label{fig:rnn}
\end{figure}

\subsection{Transformers}
As discussed in Section \ref{stm}, working memory in the human brain is a temporary storage system that manipulates and processes new information for immediate use during a task. It enables humans to focus attention on what is important. To this end, human attention has two key aspects -- selection and shifting. Through selection, the brain identifies what deserves attention, and through shifting, the brain shifts attention when important new information is introduced. Transformers \cite{vaswani2023attention}, in a way, take these ideas of working memory and attention to the realm of AI.

\begin{figure}[!h]
    \centering
    \begin{subfigure}[b]{0.45\textwidth}
        \centering
        \includegraphics[width=\textwidth]{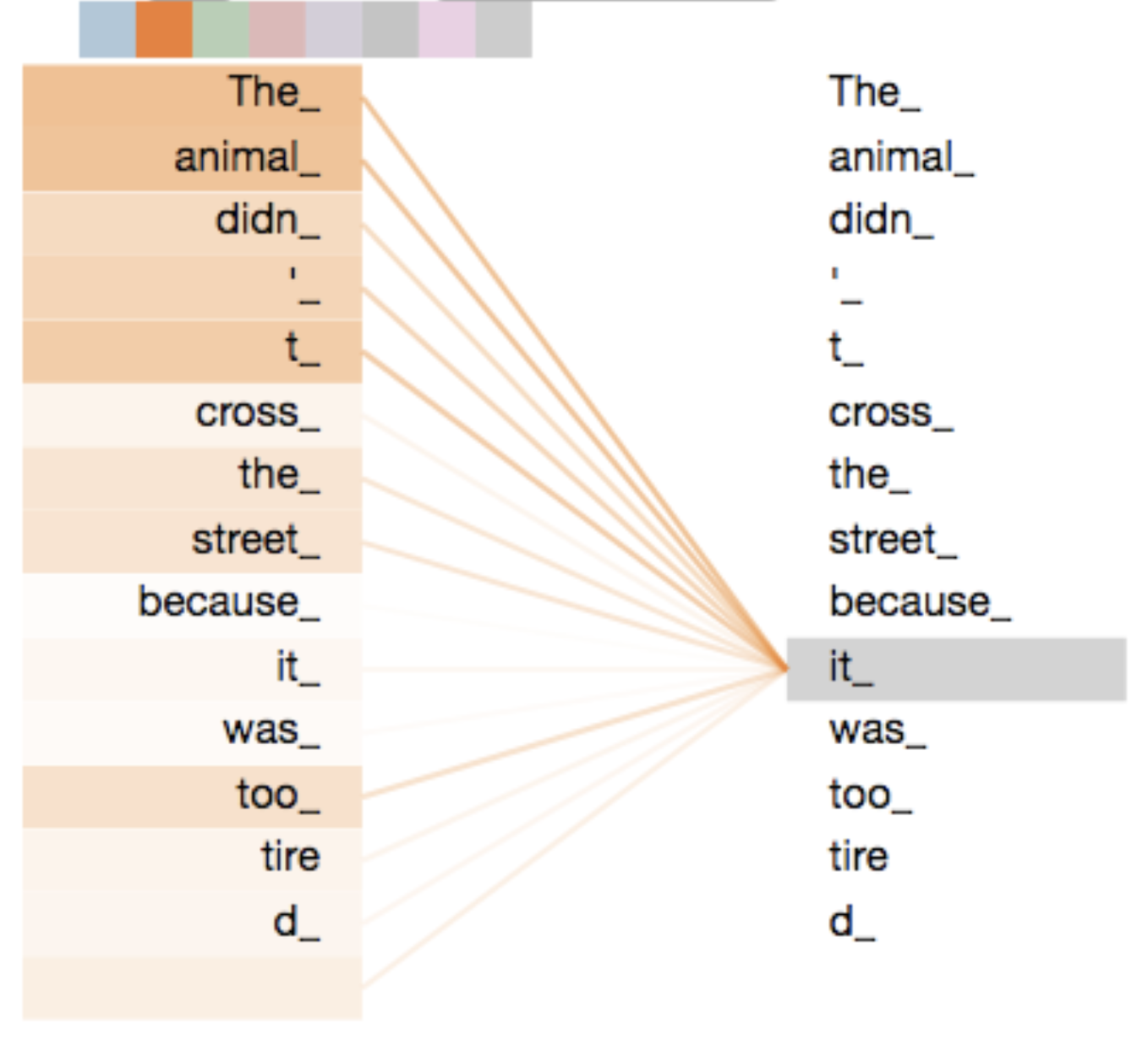}
        \caption{}
        \label{fig:self-attn}
    \end{subfigure}
    \hfill
    \begin{subfigure}[b]{0.45\textwidth}
        \centering
        \includegraphics[width=\textwidth]{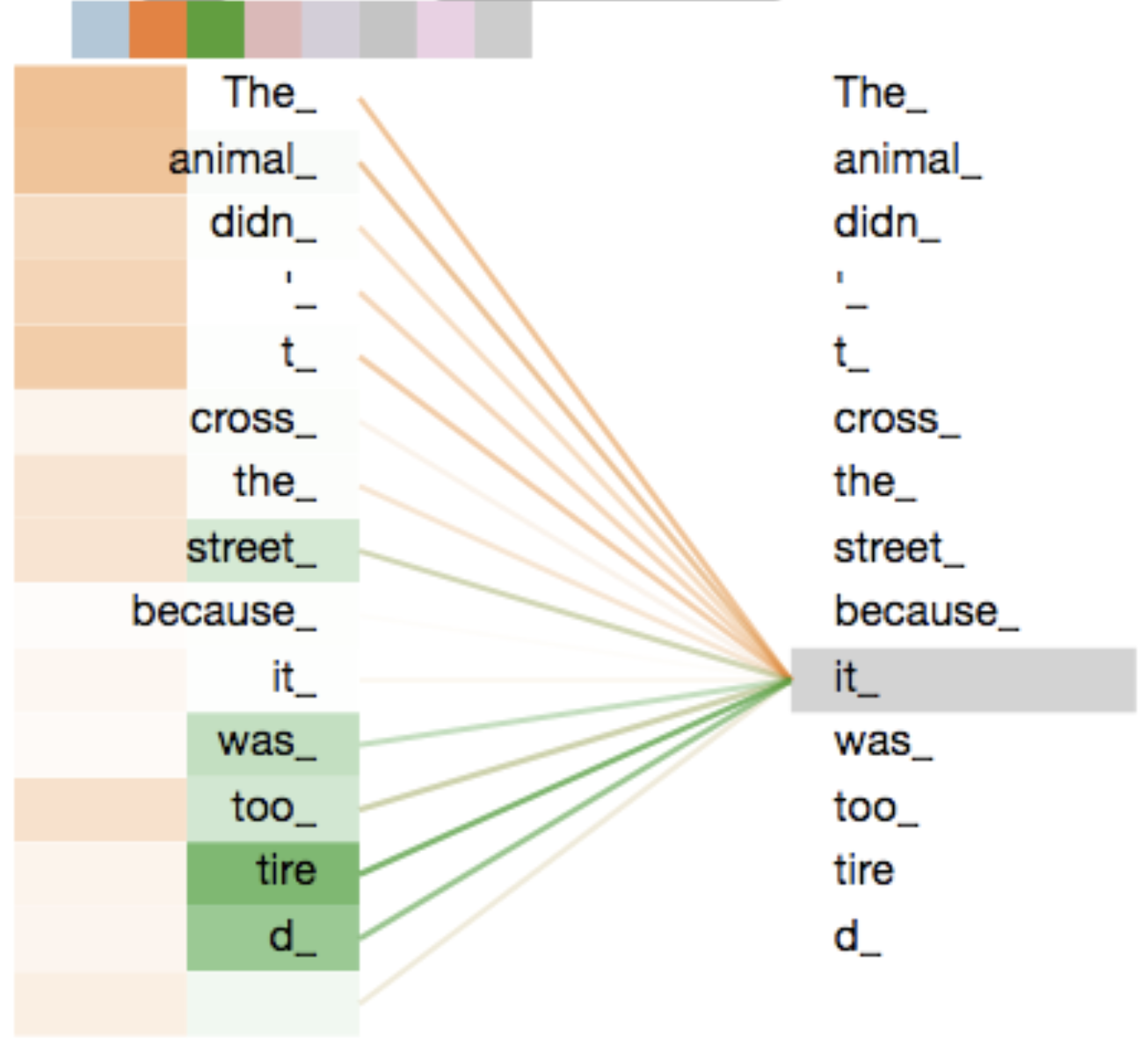}
        \caption{}
        \label{fig:milti-attn}
    \end{subfigure}
    \caption{(a) Self-attention allows Transformers to associate \texttt{it} with \texttt{animal} and (b) Multi-headed self-attention equips Transformers with the ability to shift its focus to different parts of the sentence based on the context. Figure and example are taken from \href{https://jalammar.github.io/illustrated-transformer/}{The Illustrated Transformer by Jay Alammar}}
    \label{fig:whole}
\end{figure}

To understand this, consider the example of machine translation. Consider the sentence, "\texttt{The animal didn't cross the street because it was too tired.}" When translating this sentence, the model must know what "\texttt{it}" refers to -- the animal or the street. When the Transformer is processing "\texttt{it}", self-attention allows it to associate "\texttt{it}" with "\texttt{animal}" as shown in Figure \ref{fig:self-attn}. This is AI-version of the "selection" property of the human brain.

Further, multi-headed self-attention expands Transformer's ability to focus on different parts of the input. As we encode the word "\texttt{it}", one attention head focuses most on "\texttt{the animal}", while another is focusing on "\texttt{tired}." In a sense, this means that the model's representation of the word "\texttt{it}" bakes in some of the representation of both "\texttt{animal}" and "\texttt{tired}", and based on the context the model can tell what "\texttt{it}" refers to and what is the state of "\texttt{it}." This is analogous to the “shifting” property of the human attention.

\subsection{Hopfield Networks}
\begin{figure}[!h]
    \centering
    \includegraphics[width=.27\linewidth]{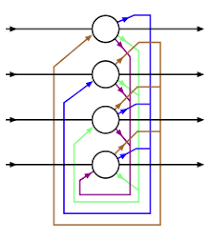}
    \caption{A Hopfield network with four units. Figure taken from \href{https://en.wikipedia.org/wiki/Hopfield_network}{Hopfield network, Wikipedia}}
    \label{fig:hopfield-network}
\end{figure}

The Hopfield network \cite{hopfield1982neural} is a type of autoassociative memory. It consists of a single layer of interconnected neurons with symmetric weights. As shown in Figure \ref{fig:hopfield-network}, it is a recurrent network, i.e. its outputs are fed back into its inputs.

\textbf{Storage Algorithm:} The Hopfield network stores information in the weights of the network. The storage algorithm for the network is given in equations \ref{eq:hopfield}. In particular, $W$ is built from the correlations between all pairs of data vector to be learned $x$.

\begin{equation}
    W_{ij} = \sum_{s=0}^{M-1} x_{i}^{s} \cdot x_{j}^{s} ] \quad \text{where } i \neq j \text{ and } W_{ij} = 0 \text{ where } i = j
    \label{eq:hopfield}
\end{equation}

The Hopfield network learns these corelations using Hebbian learning, wherein the idea is to strengthen correlated synapses and weakens negatively correlated ones. This method of learning is computationally more efficient compared to methods like backpropagation.

\textbf{Recall Process:} Recall is an iterative process where the network updates its state until it stabilizes. Equation \ref{eq:hopfield-recall} describes the update rule.

\begin{equation}
    x_i(t + 1) = f_h[\sum_{i=0}^{N-1} W_{ij} \cdot x_j(t)]
    \label{eq:hopfield-recall}
\end{equation}

Essentially, the network behaves like a hill-climbing local search algorithm and searches an energy landscape with the stored patterns as the minima. The network is guaranteed to converge to a solution, but not necessarily a stored pattern due to local optima.

\textbf{Limitations: } Hopfield networks have limited storage capacity, roughly $0.15N$ items (where $N$ = number of neurons) before recall errors become significant. This limitation is due to noise and spurious states in the network caused by crosstalk between patterns.

\subsection{Correlation Matrix Memories (CMMs)}
Correlation Matrix Memories \cite{kohonen1972cmm} is an example of a heteroassociative memory (although it can also function as an autoassociative memory).
It is a single-layer neural network with fully connected input and output neurons, forming a weight matrix $W$. This size of $W$ equals the product of input and output vector lengths.
CMM can store many pairs of input and output data vectors and they offer properties like fault tolerance and generalization.

\textbf{Storage: } The correlations between input and output pairs is stored in the weight matrix $W$ of the network. $W$ can be learned using methods like pseudo-inverse, gradient descent, and Hebbian learning.

\begin{itemize}
    \item Pseudoinverse requires complete input-output pairs and recalculates the entire weight matrix when adding a new pair, making it impractical.
    \item Gradient descent adjusts the weight matrix upon presenting new pairs but avoids the complete recalculation issue.
    \item Hebbian learning is commonly used, especially when inputs are orthogonal, as it only requires local updates and enables quick learning.
\end{itemize}

\textbf{Recall: } Recall operation is described in equation \ref{eq:cmm-recall}. It involves multiplying the input vector $x_i$ with the weight matrix $W$, creating an activity pattern on the output neurons. Often, an activation function is need to map this pattern to the final output $y_i$.

\begin{equation}
    y_i = f (W \cdot x_i)
    \label{eq:cmm-recall}
\end{equation}

\textbf{Strengths and Limitations: } 
The following are some of the advantages of using CMMs as memory:
\begin{itemize}
    \item Fault tolerance: CMMs exhibit robustness to noisy input. It can even perform recall for incomplete inputs. 
    \item Generalization: CMMs are capable of generalizing to unseen inputs.
    \item Storage capacity: Compared to Hopfield networks, CMMs have a relatively large storage capacity.
\end{itemize}
The main disadvantage of using CMMs as memory is \emph{crosstalk} between the stored patterns. As the data is stored in a decentralized fashion, when we retrieve information from the memory, the resulting activity is a mixture of the intended data and other stored data.

\subsection{Neural Turing Machines (NTMs) \cite{graves2014neural}}
\begin{figure}[!h]
    \centering
    \includegraphics[width=.5\linewidth]{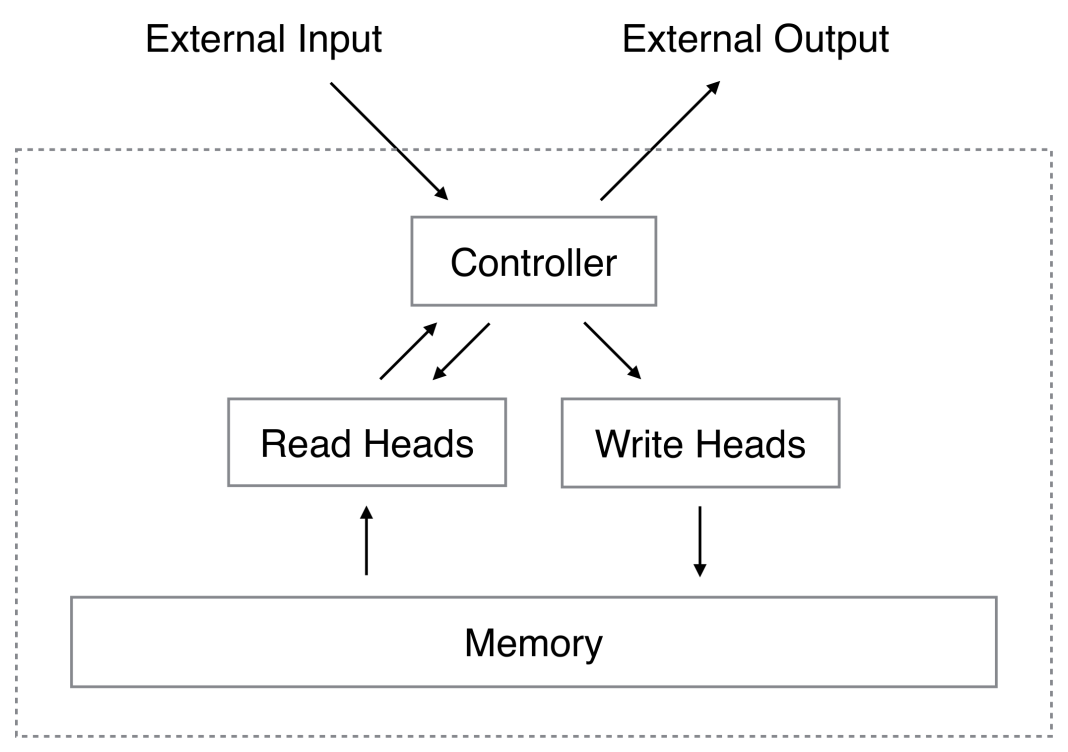}
    \caption{The NTM's controller network interacts with external inputs, generating outputs and accessing a memory matrix through read and write heads during each update cycle \cite{graves2014neural}}
    \label{fig:ntm}
\end{figure}

An NTM comprises two fundamental components: a neural network controller and an external memory bank. The controller operates similar to a conventional neural network, processing inputs and producing outputs. However, what sets NTMs apart is their ability to interact with an external memory matrix, enabling reading from and writing to this memory. This feature endows NTMs with a dynamic memory structure, allowing them to learn and store information efficiently. Fig \ref{fig:ntm} gives an overview of the NTM architecture.

\textbf{Storage: } To store information over extended time steps, an NTM maintains a memory matrix $M \in \mathbb{R}^{d_N \times d_M}$, where $d_N$ is the number of memory locations and $d_M$ is the size of each location. When new information needs to be stored, the controller determines the location and content to be written into the memory matrix. This operation involves a series of attention mechanisms, allowing the NTM to focus on specific memory locations, potentially overwriting existing data while preserving essential information. In particular, at time step $t$, the write head gives an attention vector $w_t \in \mathbb{R}^{d_N}$, which is used to weigh an \textit{erase} vector $e_t \in \mathbb{R}^{d_M}$ and an \textit{add} vector $a_t \in \mathbb{R}^{d_M}$. Eq \ref{eq:ntm-write} gives the operation performed at each memory location $M(i)$.

\begin{equation}
    \mathbf{M}_t(i) \xleftarrow{} \mathbf{M}_{t-1}(i)[\mathbf{1} - w_t(i)\mathbf{e}_t + w_t(i)\mathbf{a}_t]
    \label{eq:ntm-write}
\end{equation}

\textbf{Recall: } Recall is implemented using a differential read operation. The read head emits an attention vector $w_t \in \mathbb{R}^{d_N}$, which is used to read from the memory using Eq \ref{eq:ntm-read}.

\begin{equation}
    \mathbf{r}_t \xleftarrow{} \sum_i w_t(i)\mathbf{M}_{t-1}(i)
    \label{eq:ntm-read}
\end{equation}

\textbf{Strengths and Limitations: } NTMs exhibit several advantages over conventional neural networks. Their dynamic memory structure enables them to excel in tasks requiring sequential reasoning, algorithmic manipulation, and handling complex datasets. Moreover, their ability to read from and write to memory offers a significant advantage in scenarios where retaining and recalling past information is crucial.

However, NTMs also pose challenges. The complexity of managing an external memory structure adds computational overhead, making training and implementation more resource-intensive. Moreover, designing efficient algorithms for memory interactions and optimizing their performance remains an ongoing research area.


\subsection{Memformer}
Memformer \cite{wu2022memformer} implements an external dynamic memory for encoding and retrieving past information so as to achieve linear time complexity and constant memory space complexity for processing long sequences. In this way, it addresses efficiency issues faced by transformers that require storing all history token-level representations.

\begin{figure}[!h]
    \centering
    \begin{subfigure}{.5\textwidth}
        \centering
        \includegraphics[width=.9\linewidth]{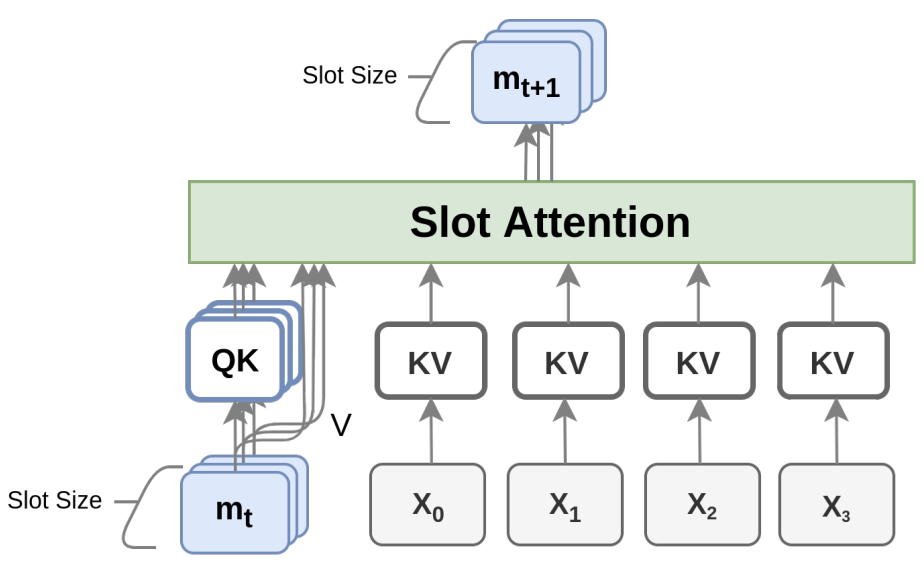}
        \caption{Memory writing using slot attention}
        \label{fig:memformer-write}
    \end{subfigure}%
    \begin{subfigure}{.5\textwidth}
        \centering
        \includegraphics[width=.9\linewidth]{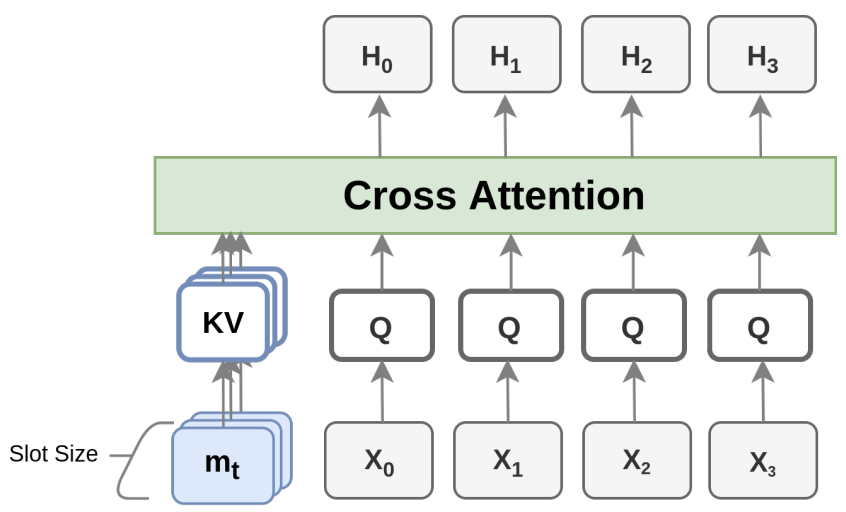}
        \caption{Memory reading using cross attention}
        \label{fig:memformer-read}
    \end{subfigure}
    \caption{Storage and recall mechanisms in Memformer \cite{wu2022memformer}}
    \label{fig:test}
\end{figure}

\textbf{Storage: } Memformer utilizes an external memory to encode and retain important information through timesteps. Memory writing and updating occur in the last layer of the encoder, allowing for storage of high-level contextual representations. In particular, it uses a slot attention mechanism to write to the memory. The idea behind slot attention is that each memory slot attends to the input sequence and to itself to generate an updated memory (as shown in figure \ref{fig:memformer-write}). Further, a forgetting mechanism to clean up irrelavant or redundant information from the memory.

\textbf{Recall: } Memory reading involves cross attention, where input sequences' queries attend over memory slots' key-value pairs to output the final hidden states. Figure \ref{fig:memformer-read} illustrates this process. The process occurs multiple times as each encoder layer incorporates a memory reading module to successfully retrieve necessary information.

\textbf{Strengths: } Experimental results demonstrate that Memformer achieves comparable performance to baselines but uses significantly less memory space (8.1x less) and is faster in inference (3.2x) for sequence modelling. The model's memory writing mechanism allows for selective update or preservation of information in memory slots, enhancing its adaptability.

\subsection{Neural Attention Memory (NAM)}
Neural Attention Memory \cite{nam2023neural} reimagines the attention mechanism as a memory architecture for neural networks. It draws inspiration from linear transformers \cite{katharopoulos2020transformers} and involves writing a memory matrix using key-value pairs and reading it with a query vector.

\textbf{Storage and recall: } NAM operates by performing matrix-vector multiplication with a memory matrix $M \in \mathbb{R}^{d_v \times d_k}$, where $d_v$ and $d_k$ are dimensions of value and key vectors respectively. Using query $q$, key $k$, and value $v$ vectors, it proposes 3 differentiable operations to interact with $M$, which are \emph{read}, \emph{write}, and \emph{erase}. 

Equation \ref{eq:nam-read} defines the read operation. 

\begin{equation}
    r = RD(M, q, p_r) = p_rMq
    \label{eq:nam-read}
\end{equation}

Here, $q \in \mathbb{R}^{d_v}$ is a unit query vector and $p_r \in [0, 1]$ is the read probability.

Equation \ref{eq:nam-write} defines NAM's write operation. It works by adding the outer product of the unit key vector $k$ and value vector $v$ to the memory $M$. Further, it also provides the mechanism to erase the current information stored in $M$ corresponding to $k$. Similar to the read operation, write and erase are regulated by write probability $p_w \in [0, 1]$ and erase probability $p_e \in [0, 1]$, respectively.

\begin{equation}
    w = WR(M, k, v, p_w, p_e) = M + p_wvk^T - p_eMkk^T
    \label{eq:nam-write}
\end{equation}

This write operation guarantees that reading with the same key $k$ yields the most recently written value $v$.

\textbf{Advantages: } The following are some strengths of NAM:
\begin{itemize}
    \item NAM offers an efficient alternative to traditional transformers, particularly addressing the quadratic bottleneck issue by using memory.
    \item Experimental results demonstrate its effectiveness in long-range tasks and compositional generalization tasks.
    \item NAM's read and write primitives can be used to implement LSTMs \cite{hochreiter1997lstm}, Neural Turing Machines \cite{graves2014neural}, and efficient Transformers \cite{katharopoulos2020transformers}.
\end{itemize}
\section{Applications and Advancements of Memory-Augmented Neural Networks}

Modern neural networks have enormous size to encode a huge amount of information in the long-term memory, i.e., their parameters. However, recent research shows that regardless of the model size, short-term memory is still important to produce interpretable, truthful and trustworthy output in a parameter-efficient manner~\citep{mialon2023augmented}. Short-term memory in neural networks is often implemented by retrieval augmentation. Analogous to the usage of short-term memory, the retrieved information is temporarily stored and discarded after the application. With the development of semi-supervised learning, neural networks are capable of getting semantically meaningful representations, and the memory is often implemented in a dense manner. Namely, the retriever compare similarities between dense queries and dense document vectors obtained from neural networks~\citep{asai2021one}. Here, we introduce the recent applications of retrieval augmentation in areas including natural language processing, computer vision and multimodal learning.

\begin{figure}[!h]
    \centering
    \includegraphics[width=0.8\linewidth]{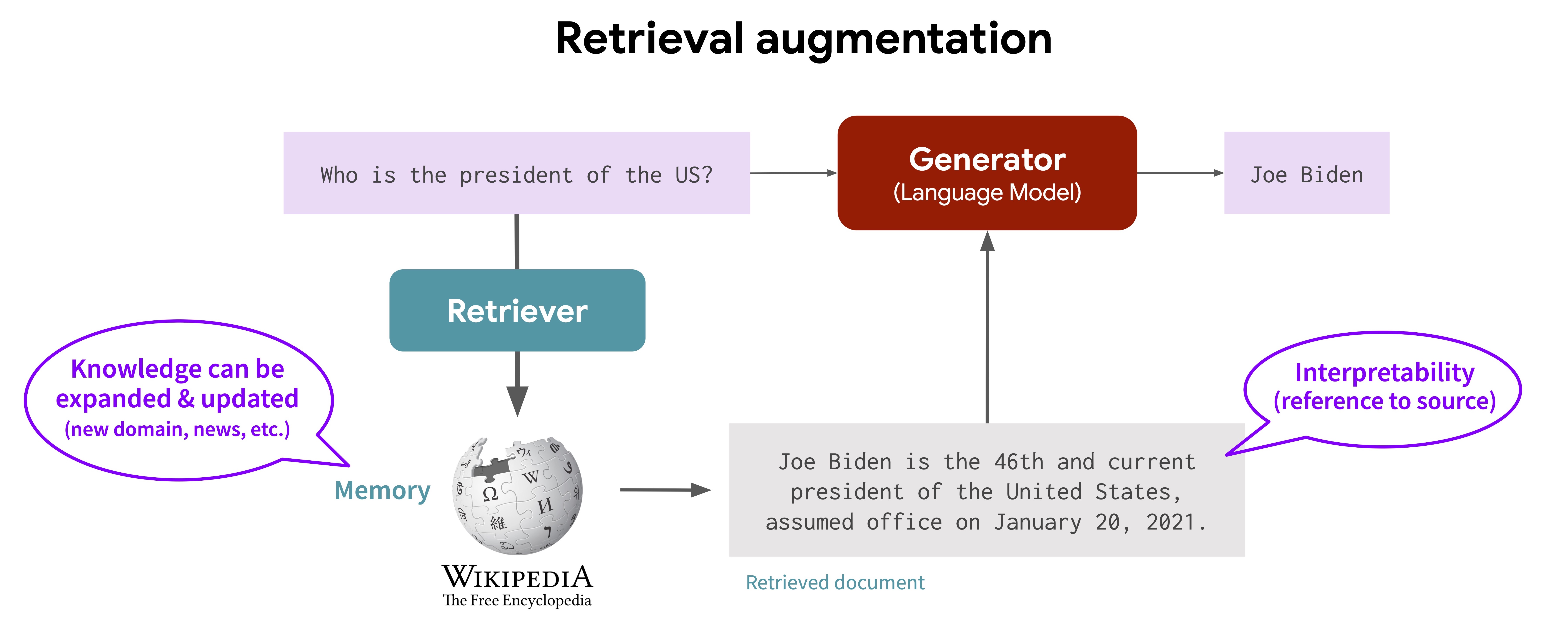}
    \caption{A typical pipeline of retrieval-augmented language models~\citep{yasunaga2023retrievalaugmented}. Given an input query, the retriever first performs a dense retrieval on an external knowledge base, then the generator produces the output conditioned on both the input query and the retrieval results.}
    \label{fig:retrieval_lm}
\end{figure}

\subsection{Natural Language Processing}
Retrieval augmentation proves to be highly beneficial in various knowledge-intensive NLP applications, especially those where factuality is a crucial requirement. This approach enriches the applications by providing access to a broader range of factual information, enhancing their accuracy and reliability. Retrieval Augmented Language Model (REALM)~\cite{guu2020realm} is the first method to jointly train a knowledge retriever and a knowledge-augmented language encoder in an unsupervised manner. Retrieval augmented generation (RAG)~\cite{lewis2021retrievalaugmented} fine-tunes a pre-trained retriever (e.g., DPR~\cite{karpukhin2020dense}) and a pre-trained sequence-to-sequence model (e.g., BART~\cite{lewis2019bart}). RAG achieves superior performance on various knowledge-intensive tasks, including question answering, question generation and fact verification. Retrieval Augmented Translation (RAT)~\citep{Hoang2023} improves neural machine translation by treating the external knowledge base as a dictionary. More recently, Chain-of-Noting (CoN)~\citep{yu2023chainofnote} proposes to enhance the robustness of retrieval augmented models by first generating sequential reading notes based on the retrieved documents, then formulating the final answer. This approach enables a comprehensive evaluation of the relevance and factuality of the retrieval results. 

Beyond the benefit of providing truthful information for knowledge-intensive tasks, retrieval augmentation also enhances the capability of language models without increasing trainable parameters. Retrieval Enhanced Transformers (RETRO)~\cite{borgeaud2022improving} augments a language model with an external knowledge base consisting of 2 trillion tokens, and achieves performance comparable to GPT-3~\cite{brown2020language} and Jurassic-1~\cite{J1WhitePaper}, which has 25x more parameters. Similarly, Atlas~\cite{izacard2022atlas} achieves state-of-the-art performance the few-shot learning capabilities in language models while still maintains a relatively small parameter size. In-Context Retrieval Augmented Language Modeling (RALM)~\citep{ram2023context} further reduces the training cost by directly prepending retrieved documents to the input and freezing the weights of language models. 

\subsection{Computer Vision}
While the ability of referring to external knowledge is naturally desirable in text-based tasks, its application in the vision domain is less common. However, the vision community has recently shown growing interest in retrieval augmentation, especially for problems dealing with low-resource domains, where learning from scarce information poses a challenge. Retrieval Augmented Classification (RAC)~\cite{long2022retrieval} tackles the problem of long-tail classification by augmenting a standard image encoder with a retrieval module. Instead of using extensive external knowledge as in NLP, RAC only uses its training set as the memory. This strategy has been effective, particularly in achieving high accuracy in tail classes, thus validating the utility of external memory in such applications. Retrieval-Augmented Customization (REACT)~\citep{liu2023learning} uses CLIP model~\cite{radford2021learning} as the retriever for relevant image-text pairs from the web-scale CLIP pretrained data. With the retrieval result, they train modualized blocks of vision transformers to perform a wide range of tasks, including  classification, detection and semantic segmentation.

Beyond classification, retrieval augmentation has also shown to be beneficial for generation tasks, enhancing the efficiency and trustworthiness. KNN-Diffusion~\cite{sheynin2022knndiffusion} tackles the problem of text-to-image generation in low-resource domains. It significantly reduces the burden of large-scale training by first retrieving relevant features encoded in the memory, then performing text-driven semantic manipulations to generate new images. Semi-Parametric Neural Image Synthesis (SPNIS)~\cite{blattmann2022semiparametric} shares an similar idea with KNN-Diffusion, with the main difference being that SPNIS only stores image representations in the external database. Retrieval-Augmented Diffusion Models~\citep{blattmann2022retrieval} underscores that integrating a retrieval module is more computationally efficient and scalable than simply increasing model complexity. This approach also has an additional advantage which is flexibility, as the augmented diffusion model can be readily adapted to new domains with changes in the knowledge base.

\subsection{Multimodal Learning}
Similar to the applications in previous sections, multimodal problems also benifit from retrieval augmentation. One of the most common applications is in Visual Question Answering (VQA). Multimodal Retrieval-Augmented Transformer (MuRAG)~\cite{chen2022murag} accesses an external multimodal memory to enhance the capabilities of language generation when dealing with VQA problems. They first use a dense retriever to retrieve the most relevant multimodal documents, then perform modality fusion by a T5 model~\citep{raffel2020exploring} to augment generation. Instead of using an off-the-shelf retriever such as CLIP or DPR, Retrieval-Augmented Visual Question Answering (RA-VQA)~\citep{lin2022retrieval} trains a multimodal retriever with the answer generator in an end-to-end manner to tackle the problem of Outside-Knowledge Visual Question Answering (OK-VQA). This neural-retrieval-in-the-loop approach improves the relevance of the retrieval results and the answer.

Retrieval-Augmented CM3 (RA-CM3)~\cite{yasunaga2023retrievalaugmented} is the first model that can retrieve and generate both text and image. It uses a pretrained CLIP model as a mutlimodal retriever and trains a CM3 Transformer~\cite{aghajanyan2022cm3} as the generator. Multimodal retrieval allows the model to access image-text pairs, enabling it to faithfully capture the visual characteristics of specific entities, such as national flags and landmarks. Retrieval Augmented Visual Language Model (REVEAL)~\citep{hu2023reveal} jointly pretrains an encoder, a retriever and a generator to perform vision-language tasks, including visual question answering and image captioning, based on multimodal knowledge. The encoder first encodes large-scale multimodal knowledge into the memory, then a retriever retrieves the most relevant knowledge from the memory, and the generator fuses the retrieval results with the input query to produce the output.

In summary, retrieval augmentation enhances the capability of neural networks in terms of both efficiency and trustworthiness. With the help of an external memory, neural networks no longer have the training overhead of encoding all information in the parameters, which enables small retrieval-augmented model to achieve similar or even better performance compared with their larger counterparts without retrieval augmentation. Moreover, the external memory provides neural networks with more reliable resources to ground on in order to provide more factual and up-to-date responses. Users can also interpret the model output more easily as they can tell which resource the model refers to.
\section{Discussion}

Having explored the numerous advantages of Retrieval-Augmented Models (RAM) in enhancing various applications, it is equally important to acknowledge and examine their limitations. In this section, we delve into the pitfalls of current RAMs, highlighting the necessity for further research to overcome these challenges.

\paragraph{Trustworthiness} Recent research has identified that the retrieval results may not be factual or relevant to the actual queries~\citep{lin2022retrieval, yu2023chainofnote}. While some solutions focus on training more selective retrievers, this approach does not entirely address the underlying problem, which stems from the presence of irrelevant or potentially harmful information in the knowledge base. This situation underscores the necessity for more effective design and filtering strategies for the external knowledge base, ensuring that the retrieved information is both accurate and pertinent.

\paragraph{Customizability} In current RAM, the retrieval process typically involves straightforward algebraic operations such as dot product or cosine similarities between the input query embeddings and document embeddings. Despite the simplicity of the retrieval process, it may not always align with specific retrieval needs. For instance, when perfomring cross-modality retrieval, cosine similarity may be an inadequate measure due to modality gaps~\citep{liang2022mind}. This issue becomes more pronounced with semantically sophisticated modalities like videos. Therefore, there is a growing need for more nuanced retrieval methods tailored to specific objectives. A pioneering approach in this direction is the use of instruction-tuned retrieval models~\citep{asai2022taskaware}, which couples the query with a retrieval instruction and demonstrates promising results.

\paragraph{Integration} Different from retrieval augmentation, an alternative approach to obtain knowledge for models is retraining or finetuning. This method has the advantage of integrating new knowledge more seamlessly with the existing information encoded in the parameters of networks. It remains unclear whether retrieval augmentation and conditioning are as effective as directly encoding the information through training. In addition, a significant drawback of the retrieval approach is computational inefficiency for inference, particularly when a large volume of queries necessitates repeatedly searching for the same results. This inefficiency highlights a need for balancing the training overhead of finetuning models and inference overhead of retrieval augmentation.

\bibliographystyle{unsrt}  
\bibliography{references}

\end{document}